\documentclass[a4paper,conference]{IEEEtran}
\usepackage[table]{xcolor}

\usepackage{comment}
\usepackage{times}
\usepackage{booktabs}
\usepackage{array}
\usepackage{booktabs}
\usepackage{multirow}
\usepackage{amssymb}
\usepackage{xcolor}
\usepackage{physics}
\usepackage{subcaption}
\usepackage{adjustbox}

\usepackage{bm}
\usepackage{url}
\definecolor{Lightgray}{rgb}{0.9176, 0.95, 0.95686} 
\DeclareMathOperator*{\argmax}{arg\,max}
\ifCLASSINFOpdf
\else
\fi
\begin{document}
%
\title{Enhancing Handwritten Text Recognition with N-gram sequence decomposition and Multitask Learning}

\author{\IEEEauthorblockN{Vasiliki Tassopoulou}
\IEEEauthorblockA{School of Electrical and\\Computer Engineering\\
National Technical University of Athens\\
Email: tassopoulouvasiliki@gmail.com}
\and
\IEEEauthorblockN{George Retsinas}
\IEEEauthorblockA{School of Electrical and\\Computer Engineering\\
National Technical University of Athens\\
Email:gretsinas@central.ntua.gr}
\and
\IEEEauthorblockN{Petros Maragos}
\IEEEauthorblockA{School of Electrical and\\Computer Engineering\\
National Technical University of Athens\\
Email: maragos@cs.ntua.gr}}


%


\maketitle

\begin{abstract}

Current state-of-the-art approaches in the field of Handwritten Text Recognition are predominately single task with unigram, character level target units. In our work, we utilize a Multi-task Learning scheme, training the model to perform decompositions of the target sequence with target units of different granularity, from fine to coarse. We consider this method as a way to utilize n-gram information, implicitly, in the training process, while the final recognition is performed using only the unigram output. 
Unigram decoding of such a multi-task approach highlights the capability of the learned internal representations, imposed by the different n-grams at the training step.
We select n-grams as our target units and we experiment from unigrams till fourgrams, namely subword level granularities. 
These multiple decompositions are learned from the network with task-specific CTC losses. 
Concerning network architectures, we propose two alternatives, namely the Hierarchical and the Block Multi-task. 
Overall, our proposed model, even though evaluated only on the unigram task, outperforms its counterpart single-task by absolute 2.52\% WER and 1.02\% CER, in the greedy decoding, without any computational overhead during inference, hinting towards successfully imposing an implicit language model.\end{abstract}


%
\IEEEpeerreviewmaketitle

\section{Introduction}
\label{sec:intro}

\begin{figure}
    \centering
    \begin{tabular}{c}
         \includegraphics[width=.85\linewidth]{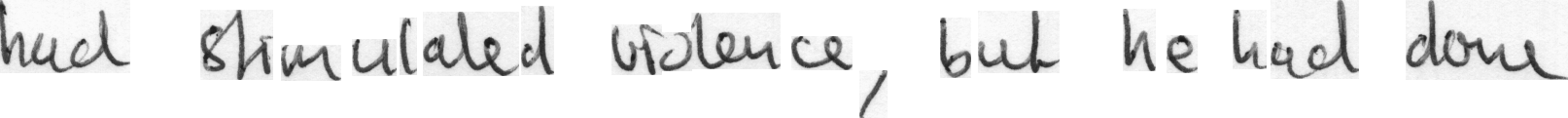} \\
         \includegraphics[width=.85\linewidth]{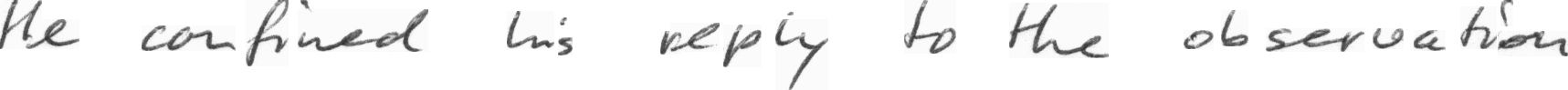} \\
         \includegraphics[width=.85\linewidth]{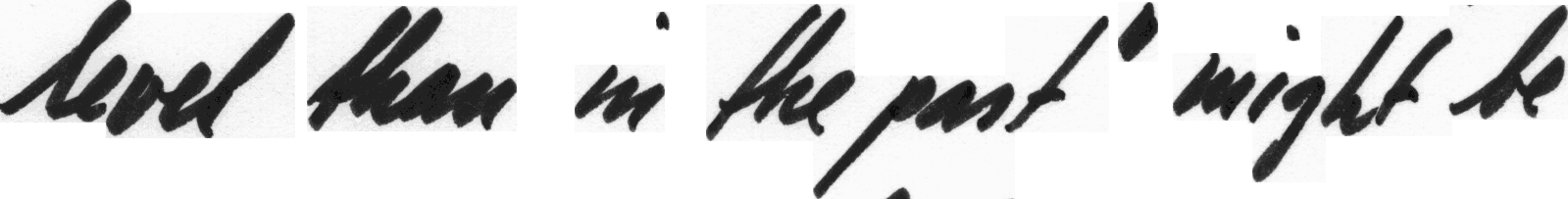}
    \end{tabular}
    \vspace{-.4cm}
    \caption{Examples of variability in cursive writing (images taken from IAM dataset~\cite{IAM})}
    \vspace{-.6cm}
    \label{fig:iam_examples}
\end{figure}

Offline Handwritten Text Recognition (HTR) is the task of digitalizing text that is depicted in an image and is a well-established task in the field of Computer Vision, while is widely considered as a challenging task due to the vast variety of writing styles (see Figure~\ref{fig:iam_examples}).
HTR is,  by its nature, a sequence transduction task. The goal is to convert a sequence of features, extracted from an image, to a sequence of text. Primarily, this was implemented with Hidden Markov Models~\cite{HMMs,hmm1,hmm2}, while contemporary deep-learning based HTR systems are relying on a class of Recurrent Neural Networks (RNNs), the Long Short Term Memory (LSTMs) ~\cite{LSTM} ~\cite{Puig, Voigt,Doe}.
A well-known paradox, know as Sarye's Paradox supports that a cursive written word cannot be recognized unless has been segmented and cannot be segmented unless is recognized. 
This paradox has been addressed by the Connectionist Temporal Classification (CTC) Algorithm \cite{CTC}, a Dynamic Programming algorithm, that does not take into account the exact segmentation of the characters, rather their possible alignments.It cares about the monotonical ordering of the targets that are depicted on the image. 
CTC algorithm has been extensively used on the HTR research and can be utilized for training Deep Learning models since it can be formulated as a differentiable loss function for end-to-end training.

According to the literature norm~\cite{Puig,Doe,Voigt}, CTC decomposes the target sequence into unigram character-level target units and subsequently, at the decoding process, the transcribed text is going to be synthesized by them. 
Following this reasoning, any sequence of characters can be formed by the decoding process, even if it does not correspond to a valid word.   
Such unconstrained decoding can be a potentially strong disadvantage and, thus, it is widely common to enhance the decoding step by fusing language knowledge from an external source. 
Specifically, statistical Language Models (LMs) are employed, by utilizing the occurring frequencies of character- and word-level n-grams as priors in the decoding process.
In other words, LMs penalize scarcely found sub-sequences of characters that may correspond to artifacts of the visual recognition step. There are several research direction that investigate decoding \cite{harald}. 

According to the aforementioned analysis, it is evident that language information is critical for building a well-performing HTR system. 
Researchers utilize character level, word level or even a hybrid LMs (word+character) in order to further improve performance. 
Even though we also employ such statistical models in our work, we also exploit the language information in an alternative novel manner by explicitly training our HTR system to learn multiple decompositions of the target sequence over different n-gram target units. 
Each n-gram decomposition is considered as a distinct task, leading to a multi-task formulation. An additional incentive that gravitates us towards the n-gram decompositions is the fact that in many cases of handwriting, letters are merged together in a way that is hard to separate them. In such cases, the model may find it easier to localize a group of letters as a whole unit, i.e the n-gram, rather than as a set of single characters. We should note that similar ideas of sub-word learning as a multi-task problem were recently explored in the Speech Recognition domain by Sanabria \textit{et al.}~\cite{Sanabria_2018, Sanab2, hmtlother}.
Furthermore, task-agnostic solutions (either HTR/Speech)  propose to learn the best possible decomposition of units \cite{GramCTC, LatSeqDe}.

\begin{figure*}[t!]
  \centering
  \includegraphics[width=.88\linewidth]{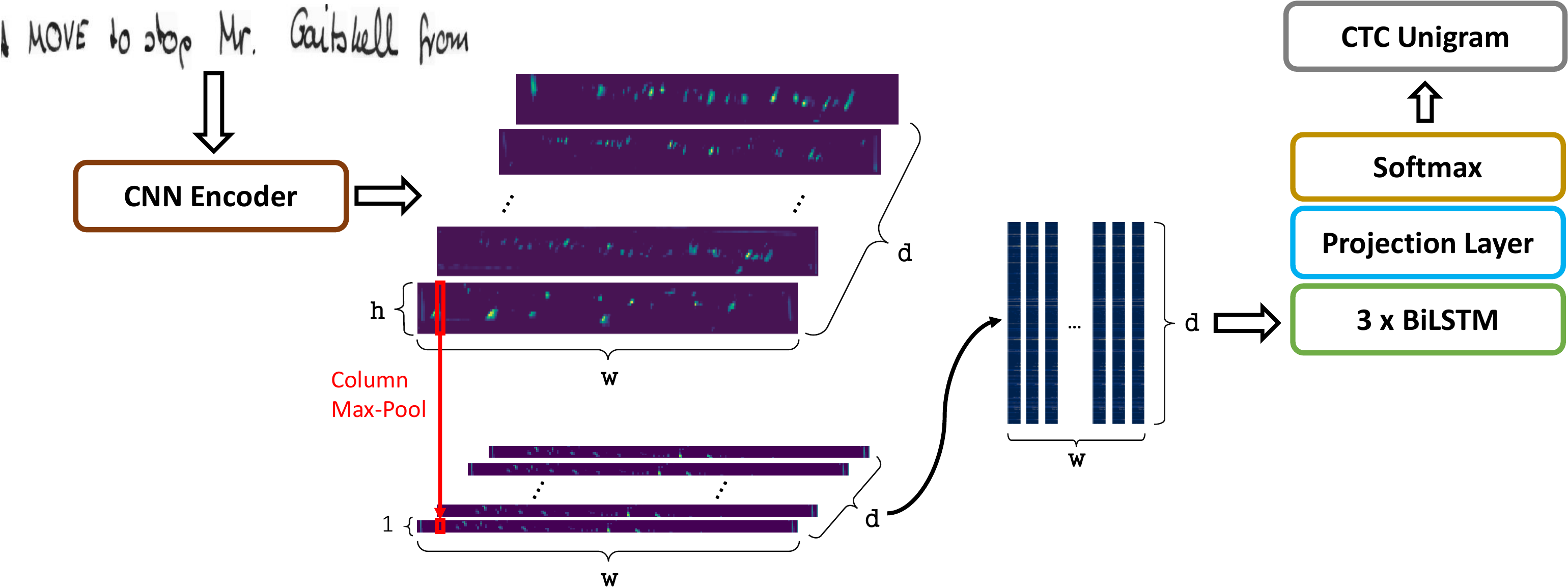}
  \vspace{-.2cm}
  \caption{\footnotesize{Baseline single-task architecture for line-level HTR, consisted of the Optical Model, the Column-wise Maxpooling, the Sequential Model and the CTC Layer.}}
  \vspace{-.5cm}
  \label{fig:baseline}
\end{figure*}

Implementation-wise, we create and evaluate a single-task (unigram only) baseline model, following the CNN + Bidirectional LSTM paradigm~\cite{Puig}, while proposing several computational efficient modifications. 
Concerning the proposed multi-task approach, using the single-task model as backbone, each level of decomposition (unigram, bigram, e.t.c.) corresponds to a different model branch which has its own CTC loss.  
The architectural choices with respect to the task-specific model branches lead to two alternatives, inspired by~\cite{Sanabria_2018}: Hierarchical Multi-task and Block Multi-task Architectures. 
Assuming a multi-task learning methodology, we aim to implicitly impose context-rich language information to intermediate generated features, previous to the final task related output. 
To this end, we evaluate the proposed approach by performing a conventional decoding process on the unigram output.
This way, we assess how multiple decompositions can enhance the internal representations of the model, while maintaining the inference time of the single-task model and preserving the simplicity of the typical decoding process.

Our paper is organized as follows, in  Section 2 we briefly describe our baseline single-task model, while in Section 3 we introduce the proposed Multi-task Architectures, capable of learning multiple decompositions, and we elaborate on our target unit selection policy. Experimental results and their discussion are presented in Sections 4, while in Section 5 conclusions are drawn. 
\vspace{-.2cm}
\section{Baseline Single-Task Model}
\vspace{-.2cm}
\label{sec:pagestyle}

\subsection{Connectionist Temporal Classification}
Connectionist Temporal Classification \cite{CTC} is a segmentation-free and alignment-free algorithm that provides a trainable loss for sequence-to-sequence transduction. It is widely used in Automatic Speech Recognition and Handwritten Text Recognition \cite{Puig, Bluche, Liu} where the goal is to convert a sequence of audio or visual features respectively into a set of target units. 

Let T be the set of characters that we want to recognize, drawn from an alphabet. The CTC algorithm demands an additional character, the blank character,  so as to separate the consecutive identical characters. Thus, it recognizes a set of T+1 characters. CTC algorithm estimates, with the use of dynamic programming, all the possible alignments, $\{a_i\}$,  that lead to the generation of the desired transcript, $y$. Specifically, CTC maximizes the probability :
\hspace{-.2cm}
\vspace{-.2cm}
\begin{equation}
 P(y|X) = \sum_{i} P(a_{i}|X)
\vspace{-.3cm}
\hspace{-.2cm}
\end{equation}
, where $a_{i}$ is an alignment that through a mapping $B$ forms $y$, i.e. $B(a_i) = y$. This mapping $B$ is also referred as the squash function, as it first concatenates all the consecutive identical characters and the removes the blank character (denoted as '-'), for example: 
$B(\text{aa-aba-bbba}) = \text{aababa}$.

\subsection{Single-task Learning}

The proposed architecture for the Single-task approach lends from popular and extensively studied architectures in the field of Handwritten Text Recognition \cite{Puig, Bluche, Liu}. 
Similarly to typical HTR architectures, our system is consisted of an Optical Model (CNN) topped by a context-aware Sequential Model (LSTM). 
However, contrary to typical flattening operations, our approach assumes a column-wise max-pooling operation to convert the optical output into a sequence of features, as requested by the Sequential Model.
The overview of the proposed single-task baseline is depicted in Figure~\ref{fig:baseline}.

We briefly describe the sub-modules of the baseline architecture: Optical Model, Column-wise Max-pooling and Sequential Model.
The \textbf{Optical Model} consists of several consecutive convolutional layers, aiming to generate discriminative visual representations. 
It takes as input the line image and generates as set of $d$ feature maps of size $h \times w$ (see Figure~\ref{fig:baseline}).   
Concerning its structure, blocks of consecutive $3\times 3$ convolutions are connected by max pooling operations of both kernel and stride equal to 2, which downsample the generated intermediate feature maps. Each convolution is followed by a batch-norm layer and a ReLU non-linearity, while each block contains several convolutional layers of equal output channels.
Overall, the employed CNN structure can be  denoted as: $2 \times \left[3\times 3, 32\right], M, 4 \times \left[3\times 3, 64\right], M, 6 \times \left[3\times 3, 128\right], M, 2 \times \left[3\times 3, 256\right]$, where $\left[k \times k, c\right]$ denotes a single convolutional layer with kernel size $k \times k$ and number of output channel $c$, while 'M' denotes a max-pooling operation.
The \textbf{Column-wise Max-pooling} operation is responsible for converting an visual output of feature maps into an appropriate sequence of features, ready to be fed into the Sequential Model. 
Typical HTR approaches, assume a column-wise approach (towards the writing direction) to ideally simulate a character by character processing. 
Flattening of the extracted feature maps would result into a sequence of length $w$ with feature vectors of size $hd$, while max-pooling results to reduced feature vectors of size $d$. 
Moreover, max-pooling generates translation invariant features compared to flattening.   
Finally, the \textbf{Sequential Model} is responsible for creating context-rich (e.g. encodings of consecutive characters) feature sequences to be fed into the CTC loss.
This model is consisted of three stacked Bidirectional LSTMs~\cite{LSTM, BiLSTM} of hidden size 256, followed by a linear projection layer and a softmax function. 
The projection creates features with size equal to the number of possible tokens/target units, while the softmax converts these feature vectors into probability distributions (e.g. the final output defines the probability of character 'a' at column $i$).    

\vspace{-.2cm}
\section{Multi-task Learning Architectures}
\vspace{-.2cm}
\label{sec:typestyle}

In this section, we will elaborate motivation and structure of two distinct architectures that can perform simultaneously multiple tasks, each represented by a different auxiliary loss term. 
In the context of our work, multi-task learning refers to explicitly learning different decompositions of the target sequence with respect to the target unit granularity.
Therefore, concerning text decomposition, using different n-grams as target-units of different scales is the straightforward solution, e.g. unigrams, bigrams e.t.c., as shown in Table~\ref{table:ngrams}. 
Note that the decomposition is performed in a sliding window fashion of unitary stride, in order to capture every possible n-gram.  
Even though the concept of including n-grams to the learning process is attractive, one major challenge is the exponentially increasing number of possible target units as we consider higher scales. 
Note that higher level tokens, e.g. at word level, may appear scarcely and thus have a minor impact to the learning process. 
To this end, we only consider the scales up to fourgrams and a subset selection process: 
every possible combination of the 26 letters is considered for bigrams (excluding the scarcely found digits and special characters), while only the top 1000 frequent trigrams and fourgrams are considered as valid. 
To this end, we include as possible targets only the most frequent n-grams, up to fourgrams.
We consider as useful n-grams those who composed from the 26 letters.
Assume subset of n-grams affects the decomposition, as presented in Table~\ref{table:ngrams}, since we may skip specific tokens. 
For example, considering the trigram case of Table~\ref{table:ngrams}, if both 'ett' and 'tte' are omitted, the resulting decomposition should be "bet-ter". 
In such a setting, blank token does not only acts as a separator between tokens, but also represents every invalid token, that cannot be classified and is omitted from the decomposition. 

\begin{table}
\centering
\begin{tabular}{l*{6}{c}r}
Target Units & Word Decomposition  \\
\hline
Unigram &  b-e-t-t-e-r  \\
All Bigrams &  be-et-tt-te-er \\
Partial Bigrams & be-tt-er  \\
All Trigrams &   bet-ett-tte-ter \\
Partial Trigrams & bet-tte-ter \\  
\end{tabular}
    \vspace{-.2cm}
\caption{\footnotesize{Example of a word decomposed into unigram, bigram and trigram target units. The character '-' denotes the blank CTC character. In case, a bigram, trigram or any other target unit of coarser granularity is missing then it is substituted by the blank character.}}
\label{table:ngrams}
\vspace{-.5cm}
\end{table}

Regarding the multi-task formulation, each scale (unigram, bigram, e.t.c.) is represented by a different auxiliary loss term and thus the overall loss is expressed as the sum of these terms, as shown in the following example for a word example and decomposing up to trigrams. If we denote as $y_u, y_b, y_c$ the posterior probabilities over the unigram, bigram ans trigram character level target units, we have: 

\hspace{-.1cm}

\begin{align*}
    L(\{y_u, y_b, y_t\}, "better") &= L^{unigrams}_{CTC}(y_u, b\text{-}e\text{-}t\text{-}t\text{-}e\text{-}r) \\
    &+ L^{bigrams}_{CTC}(y_b, be\text{-}et\text{-}te\text{-}er) \\ 
    &+ L^{trigrams}_{CTC}(y_t, bet\text{-}ett\text{-}tte\text{-}ter)
\end{align*}

 \hspace{-.1cm}
 
\begin{figure}[ht!]
    \centering
    \includegraphics[width=.68\linewidth]{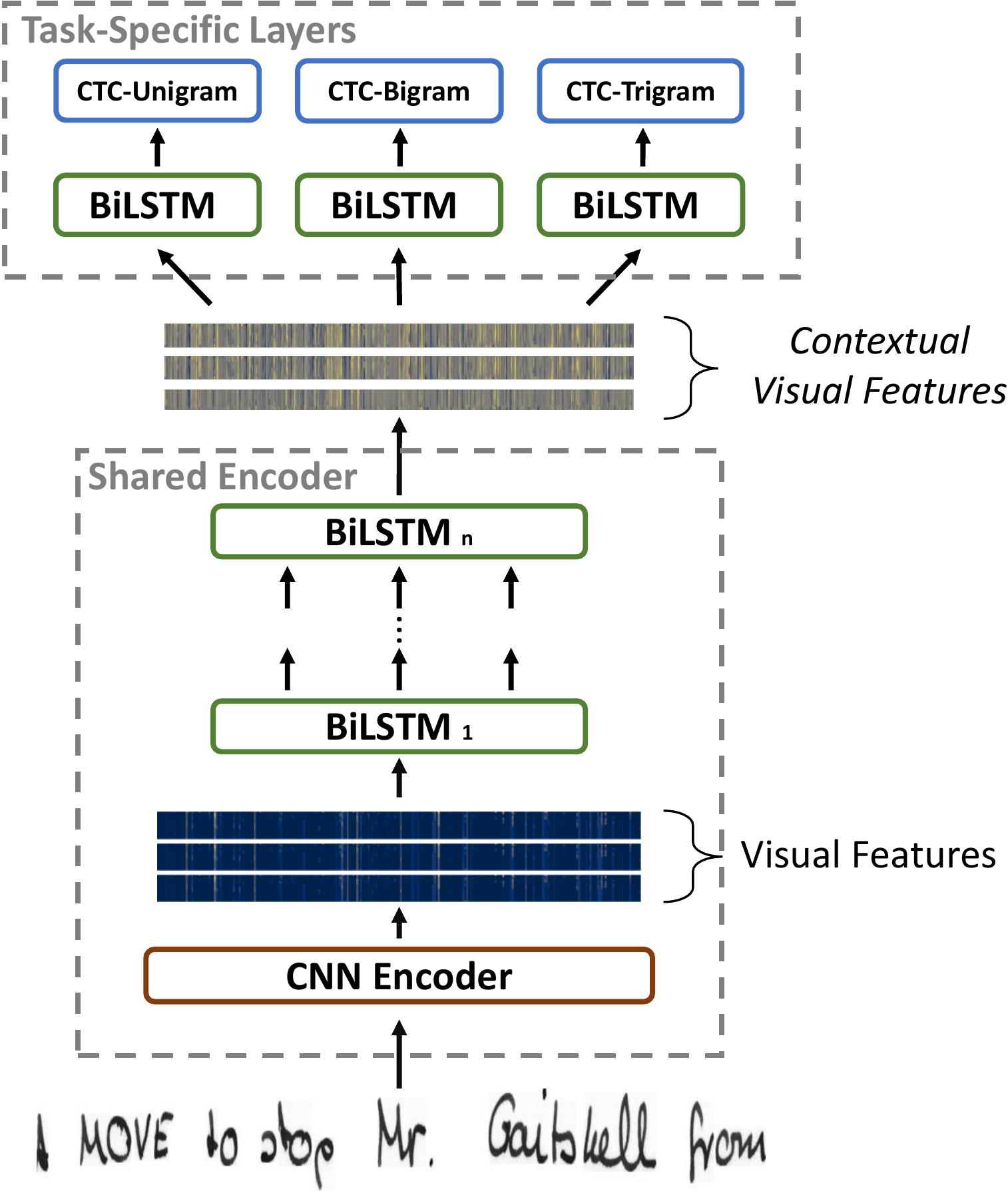}
    \vspace{-.2cm}
    \caption{\footnotesize{Architectural overview of the Block Multi-task (BMT) model for the case up to trigrams.}}
    \label{fig:bmtl}
\end{figure}

\begin{figure}[ht!]
    \centering
    \includegraphics[width=.75\linewidth]{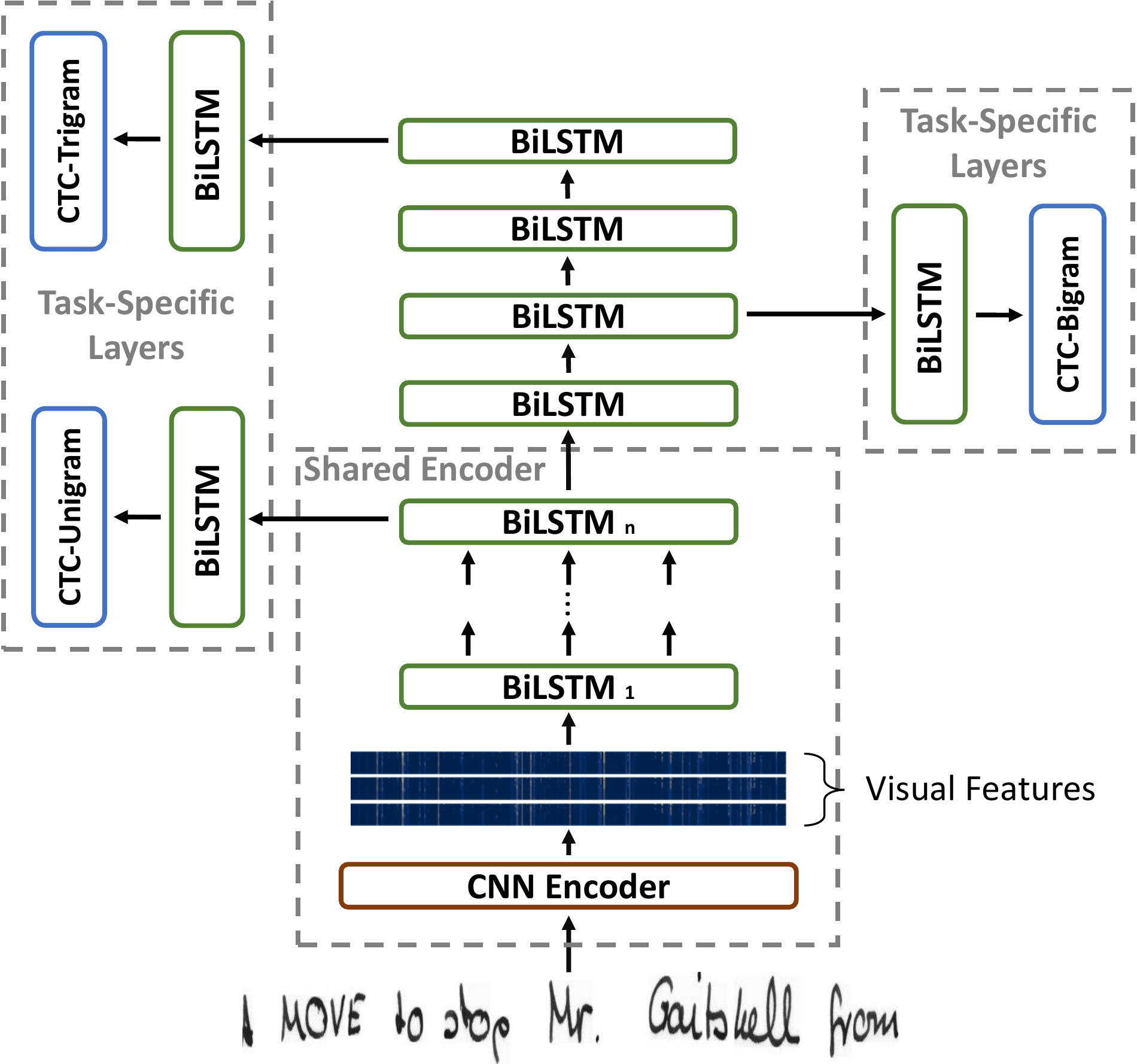}
        \vspace{-.2cm}
    \caption{\footnotesize{Architectural overview of the Hierarchical Multi-task (HMT) model for the case up to trigrams.}}
    \vspace{-.5cm}
    \label{fig:hmtl}
\end{figure}

As the loss suggests, the multi-task architectures would focus on computing the different n-gram estimations.   
The straightforward solution is to create a different branch for each scale in a parallel manner. This approach is dubbed as Block Multi-task Architecture (BMT). 
Nonetheless, higher level of n-grams correspond to more context-rich information of the previous level, hinting towards a hierarchical structure of branches, namely the Hierarchical Multi-task Architecture (HMT).
The overview of these architectures is summarized at Figures~\ref{fig:bmtl} and~\ref{fig:hmtl}, for BMT and HMT respectively, where their architectural differences are evident.
Both architectures share the same backbone as the baseline system, presented in the previous section, consisted of an optical and a sequential encoder.  
Concerning BMT architecture, after the shared encoder, a unit-specific module (one for each task) processes the hidden representations generated by the shared encoder, resulting to parallel distinct flows of information.
On the contrary, task-specific layers are built hierarchically for the case of HMT architecture, which means that between two CTC layers are inserted intermediate BiLSTM layers that learn context-rich representations, capable to describe coarser granularities.
In other words, the HMT architecture enables learning different decompositions of the target unit in different parts of the network that are built up hierarchically, one on top of the other.


\hspace{-.2cm}

Having defined the multi-task objective and possible architectures to implement them, the main question that remains concerns the prediction step.
Even though, conceptually, the existence of multiple flows of information at different scales should be beneficial, their translation into a single estimated character sequence is far from trivial.
As we have already pointed out at the introduction, our goal is not to explicitly use this extra information flow, but rather use it only on the training scheme in order to assist the generation of context-aware and highly discriminative features at intermediate shared layers of the proposed architectures.
This way, we ideally expect to improve the performance even when we evaluate only the unigram branch which can be decoded efficiently into a predicted sequence, without any complex scheme of combining information of different scales. 

\vspace{-.2cm}
\section{Decoding the Network Output}
\vspace{-.2cm}
\label{sec:exp}

As we have stated above, in order to keep the decoding procedure fast and simple we use only the branch that contains the unigram posterior probabilities. Irrespective of the architecture, we utilize two decoding algorithms. At first we apply the simple greedy decoding algorithm. If we symbolize with $P(x_{t}|\bm X)$ the posterior probability of character $x$ at time $t$ given the input image, we have : \textbf{$X$}.
\begin{align*}
    y_{dec} &= B(\argmax_x \prod_{t=1}^{T}{P(x_t| \pmb X)})
\end{align*}

The greedy decoding algorithm has the disadvantage that, in the case of most character level target units of fine coarse granularity, any word can be created, which is potentially negative as the formed word may be non-existent. Thus,  we employ CTC Beam Search algorithm, a dynamic algorithm, that allows to take into account external language information via word level and character level language model. The CTC Beam Search algorithm \cite{CTCdec} provides a means of decoding the grid of posterior probabilities by making use external language information, via a statistical language model and the dynamic programming. In this decoding algorithm whether the word exists in the corpus plays a role in the formulation of the transcription, in contrast with the greedy decoding where the algorithm bases its decision on the probabilities that we obtain from the model.

If we denote with $P_{LM}$ the external Language Model (LM) probability of $x_t$ character to be added to the already formed sequence of characters $y_{dec}$ of the previous timepoint, we have : 
\begin{align*}
    y_{dec} &= B( \argmax_x \sum (\prod_{t=1}^{T}{P(x_t| \pmb X) \cdot P_{LM}(x_t|y_{dec}(t-1))}))
\end{align*}

\vspace{-.2cm}
\section{Experimental Evaluation}
\vspace{-.2cm}
\label{sec:exp1}

\begin{figure}
    \centering
    \begin{tabular}{c}
    \includegraphics[width=.95\linewidth]{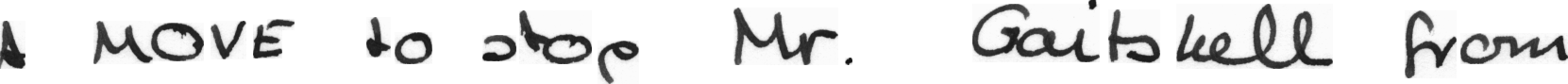} \\
    \footnotesize{(a) initial line image}\\
    \includegraphics[width=.95\linewidth]{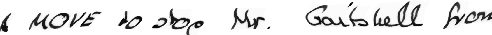} \\
    \footnotesize{(b) local affine example}\\
    \includegraphics[width=.95\linewidth]{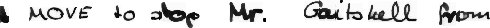} \\
    \footnotesize{(c) local morphological example}
    \end{tabular}
    \vspace{-.2cm}
    \caption{\footnotesize{Examples of random augmentations according to our local affine and morphological scheme.}}
    \vspace{-.6cm}
    \label{fig:aug_examples}
\end{figure}

The proposed augmentation scheme provides a minor, yet constant, boost in performance and therefore is considered as a default step in out pipeline.  
Figure~\ref{fig:aug_examples} contains examples of this augmentations scheme, including local affine and morphological distortions.
Training is performed over 100 epochs using RMSProp optimization. Learning rate starts from $10^{-3}$ and then is further reduced to $10^{-4}$ after half of the epochs. 

As we have already stated, decoding is performed only on the unigram posteriors.  
We consider two alternatives: greedy and Beam Search decoding ~\cite{CTCdec}. 
The later, allows us to add explicitly language information with the use of statistical language models. 
Thus, for decoding purposes we create both character level and word level language models from corpora Brown  and Lob and from the latter we have removed the IAM test set. We created the LMs with the KenLM tool~\cite{KenLM}.
The employed evaluation metrics are the Word Error Rate (WER) and the Character Error Rate (CER).
Both metrics rely on Levenshtein Distance between the predicted and the target text. 
CER corresponds to the ratio of miss-recognized characters, while WER corresponds to the ratio of miss-recognized words. 
A word is considered as falsely recognized even if a single character is off (in other words, if Levenshtein distance between two word tokens is not zero).
All the symbols (numbers, special marks) are included in the evaluation process. 

First, we consider the case of unigram-level greedy decoding. 
State-of-the-art results along with the performance of our proposed models appear in Table~\ref{table:exploration}.
Notably, our single-task architecture outperforms both related works from the literature. 
specifically, Pham et al.~\cite{Pham} utilized 2D-BiLSTMs, while Puigcerver et al.~\cite{Puig} used a similar architecture to the proposed single-task baseline, whilst having twice as many parameters compared to ours.  
Table~\ref{table:exploration} also contains the evaluation of the different multi-task variants at different granularities, using the unigram level greedy decoding scheme. 
As reported, we utilize only the unigram posteriors so as to assess the impact of the multi-task reasoning on the internal representations of the trained model.
From the results, it is evident that the use of this extra n-gram information, only on the training step, can significantly boost the model's ability to generalize and consequently the system's performance.
Indeed,  Learning in the terms of training multiple CTC decompositions of different granularity, from fine-to-coarse target units,  boosts the model's ability to generalize and thus we report improvement in the WER and CER metrics.
Moreover, as the language information increases by using higher n-gram scales we obtain better results.
We should stress that only the bigram information results to a considerable boost, while extra n-gram scale provide slight, yet consistent, performance increase.
Note that all reported variants have the same complexity, since only the unigram branch is used for evaluation.
Furthermore, no important difference among the HMT and BMT alternatives is observed, even though conceptually the former describes better the contextual nature of the n-grams. 
This can be addressed to the fact that HMT has a more complex structure which may affect the back-propagation process and hinder optimization/training, since each n-gram level straightforwardly affects its predecessor. 
Next,  we compare our unigram-level greedy decodings with the ones from previous papers. We refer to a 1D-LSTM approach \cite{Puig} as the LSTMs utilized in our architecture and one architecture \cite{Pham} that utilized 2D-LSTM. 
Summarizing, our multi-task architectures significantly outperform the majority existing approaches at the greedy decoding setting, while being on par with a very recent and computational demanding sequence to sequence approach~\cite{s2shtr}.
Specifically, our best performing variant, i.e. BMT with all n-gram scales, which will be the default network for further evaluation, has an improvement of absolute 2.52\% in the WER and 1.02\% in the CER compared to the very similar network of ~\cite{Puig}, by just utilizing the unigram-level posteriors, without any extra computational overhead.  

\begin{table}[htb]
\centering
\resizebox{.9\linewidth}{!}{
\begin{tabular}{c c c } 
\hline
 N-Grams  & WER \% & CER \% \\ [0.5ex]
  \hline
  \rowcolor{Lightgray}
\multicolumn{3}{c}{\textit{Single-Task}} \\
\hline\hline
Pham \textit{et al.} \cite{Pham} & 35.10 & 10.80 \\
Puigcerver \textit{et al.} \cite{Puig}  &  20.20 & 6.20  \\ 
Castro \textit{et al.} \cite{castro} & 24.00 & 6.64 \\
Michael \textit{et al.} \cite{s2shtr} & - & 5.24 \\ 
\hline 
1-gram (ours)  & 19.10 & 5.60\\
 \hline
  \rowcolor{Lightgray}
 \multicolumn{3}{c}{\textit{Hierarchical MT}}\\
\hline\hline 
1-grams + 2-grams & 17.72 &  5.21 \\
1-grams + 2-grams + 3-grams & 17.70 & 5.37  \\
1-grams + 2-grams + 3-grams + 4-grams & \textbf{17.68} & 5.29  \\
\hline
 \rowcolor{Lightgray}
\multicolumn{3}{c}{\textit{Block MT}} \\
\hline\hline 
1-grams + 2-grams & 17.96 &  5.28  \\
1-grams + 2-grams + 3-grams  & 17.90 & 5.30  \\
1-grams + 2-grams + 3-grams + 4-grams & \textbf{17.68} & \textbf{5.18}
\end{tabular}}
\vspace{-.2cm}
\caption{\footnotesize{The impact of different scales of n-grams at training step is explored. Results correspond to line-level greedy decoding for SoA methods along with the proposed Single-Task, HMT and BMT architectures.}}
\vspace{-.2cm}
\label{table:exploration}
\end{table}

As we have already mentioned, decoding is typically performed with the assistance of an external statistical language model.
Such LMs make the decoding process more complex, implemented by a beam search algorithm, but provide noteworthy performance increase.
In Table~\ref{table:lms}, we explore the impact of character- and word-level LMs on the performance of both the single-task and the best performing multi-task models. 
The results indicate that external LMs can operate in a complementary manner to the internal implicit LM, that multi-task learning attempts to impose.
Specifically, concerning WER, the difference in performance between single-task and multi-task models is retained, regardless the external LM applied. 
However, CER differences seem to be absorbed by the external LM.

\begin{table}[htb]
\centering
\resizebox{.65\linewidth}{!}{
\begin{tabular}{c c c} 
 \hline
 Architecture &  WER \% & CER \% \\ [0.5ex] 
  \hline
\rowcolor{Lightgray}
\multicolumn{3}{c}{\textit{CTC Greedy Decoding}} \\
\hline\hline
 Single-Task  & 19.10 & 5.60\\
 BMT & 17.68 & 5.18 \\
 \hline
\rowcolor{Lightgray}
\multicolumn{3}{c}{\textit{CTC BeamSearch 4-Gram CharLM}} \\
\hline\hline
 Single-Task & 18.14 & 5.64  \\ 
 BMT  &  16.72 &  5.28 \\
\hline
\rowcolor{Lightgray}
\multicolumn{3}{c}{\textit{CTC BeamSearch 4-Gram WordLM}} \\
\hline\hline
 Single-Task &  14.81 & 4.60  \\ 
 BMT & \textbf{13.62} &  \textbf{4.60} \\
 \end{tabular}
 }
  \vspace{-.2cm}
\caption{\footnotesize{Impact of external LMs in our Single-task and BMT architectures.}}
\vspace{-.5cm}
\label{table:lms}
\end{table}

\hspace{-.2cm}

Table~\ref{table:soalm} contains the state-of-the-art comparison of methods that use word-level LMs for line-level decoding. The majority of the methods reported utilize several techniques at the same time in order to further improve the performance of their system, including hybrid LMs~\cite{Doe, Voigt, Puig}, Lexicon~\cite{Pham, harald}, e.t.c. 
Furthermore, methods~\cite{Doe, Voigt, Puig} perform paragraph-level decoding in order to maximize the possible context and consequently enhance the impact of word-level LMs.
Contrary to high complexity architectures (e.g. 2d BiLSTMs) and high complexity decoding schemes, as the aforementioned, the proposed approach achieves recognition results very close to SoA, with minor inference and decoding overhead.  

\begin{table}[ht!]
\centering
\resizebox{.9\linewidth}{!}{
\begin{tabular}{c c c} 
 \hline
Method &  WER \% & CER \% \\ [0.5ex] 
 \hline\hline
 Pham \textit{et al.}~\cite{Pham} & 13.60 & 5.10 \\  
 Doetsch \textit{et al.}~\cite{Doe}(*) & 12.20 & 4.70 \\
Voigtlaender \textit{et al.}~\cite{Voigt}(*) & 9.30  & 3.50 \\ 
Puigcerver \textit{et al.} \cite{Puig} (*)  &  12.20 & 4.40  \\ 
Scheidl \textit{et al.}~\cite{harald} & 11.01 & 5.62 \\ 
\hline
 BMT (ours)  &  13.62 &  4.60 
\end{tabular}
}
\vspace{-.2cm}
\caption{\footnotesize{Report of SoA HTR systems, using CTC Beam Search Decoding with word-level LMs. (*) denotes paragraph-level evaluation and the corresponding references are added to the table for completeness, direct comparison with them is not fair.}}
\vspace{-.2cm}
\label{table:soalm}
\end{table}

\vspace{-.2cm}
\section{Conclusions}
\vspace{-.2cm}

In this paper, we proposed a novel way to implicitly integrate domain knowledge (language information via character level n-grams) in the HTR task. This is accomplished by training the network to learn decompositions at different n-gram granularities in a multi-task manner. Our multi-task approaches outperform the closest to us architecture \cite{Puig}, in the greedy scheme, by absolute 2.52 \% in the WER and \%1.02 in the CER and maintaining the inference time the same as the single-task model. Future endeavors for further improving the multitask scheme for HTR, include the exploration of residual architectures and  the hybrid decoding using not only unigrams but also any other branch that is in the multitask model in combination with explicit language information.  

\vspace{-.2cm}
\section*{\large Acknowledgements}
\vspace{-0.2cm}
The work of Prof. Petros Maragos was supported by the Hellenic Foundation for Research and Innovation (H.F.R.I.) under the “First Call for H.F.R.I. Research Projects to support Faculty members and Researchers and the procurement of high-cost research equipment grant” (Project: "SL-ReDu", Project Number: 2456).


\begin{thebibliography}{1}

\bibitem{Sanabria_2018}
Ramon Sanabria, Florian Metze,
\newblock ``Hierarchical Multi Task Learning With {CTC},''
\newblock in {\em Workshop on Spoken Language Technology}, 2018.

\bibitem{LSTM}
Sepp Hochreiter and J\"{u}rgen Schmidhuber,
\newblock ``Long Short-Term Memory,''
\newblock in {\em Neural Computing}, 1997.

\bibitem{CTC}
Alex Graves and Santiago Fern\'{a}ndez and Faustino Gomez and J\"{u}rgen Schmidhuber,
\newblock ``Connectionist Temporal Classification: Labelling Unsegmented Sequence Data with Recurrent Neural Networks,''
\newblock in {\em International Conference on Machine Learning}, 2006.

\bibitem{Seq2Seq}
Ilya Sutskever and Oriol Vinyals and Quoc V. Le,
\newblock ``Sequence to Sequence Learning with Neural Networks,''
\newblock in {\em Neural Information Processing Systems}, 2014

\bibitem{CTCdec}
Alex Graves and  Navdeep Jaitly
\newblock ``Towards End-to-End Speech Recognition with Recurrent Neural Networks,''
\newblock in {\em International Conference on Machine Learning}, 2014
\bibitem{Puig}
Joan Puigcerver,
\newblock ``Are Multidimensional Recurrent Layers Really Necessary for Handwritten Text Recognition?,''
\newblock in {\em International Conference on Document Analysis and Recognition}, 2017.

\bibitem{Bluche}
T. {Bluche} and R. {Messina},
\newblock ``Gated Convolutional Recurrent Neural Networks for Multilingual Handwriting Recognition,''
\newblock in {\em International Conference on Document Analysis and Recognition}, 2017.

\bibitem{Liu}
Q. {Liu} and L. {Wang} and Q. {Huo},
\newblock ``A study on effects of implicit and explicit language model information for DBLSTM-CTC based handwriting recognition,''
\newblock in {\em International Conference on Document Analysis and Recognition}, 2015.

\bibitem{BiLSTM}
Graves, Alex and Fern\'{a}ndez, Santiago and Schmidhuber, J\"{u}rgen,
\newblock ``Bidirectional LSTM Networks for Improved Phoneme Classification and Recognition,''
\newblock in {\em International conference on artificial neural networks}, 2005.

\bibitem{IAM}
Urs-Viktor Marti and Horst Bunke
\newblock ``The IAM-database: an English sentence database for offline handwriting recognition,''
\newblock in {\em International Journal on Document Analysis and Recognition}, 39-46, 2002

\bibitem{LevDist}
Frederic P. Miller and Agnes F. Vandome and John McBrewster,
\newblock ``Levenshtein Distance: Information Theory, Computer Science, String (Computer Science), String Metric, Damerau?Levenshtein Distance, Spell Checker, Hamming Distance,''
\newblock in {\em International Journal on Document Analysis and Recognition}, 39-46, 2009

\bibitem{KenLM}
Kenneth Heafield,
\newblock ``KenLM: Faster and Smaller Language Model Queries,''
\newblock in {\em Workshop on Statistical Machine Translation}, 2011

\bibitem{Pham}
Vu Pham and Christopher Kermorvant and J{\'{e}}r{\^{o}}me Louradour
\newblock ``Dropout improves Recurrent Neural Networks for Handwriting Recognition,''
\newblock in {\em International Conference on Frontiers in Handwriting Recognition} , 2014

\bibitem{Voigt}
Paul Voigtlaender and Patrick Doetsch and Hermann Ney
\newblock ``Handwriting Recognition with Large Multidimensional Long Short-Term Memory Recurrent Neural Networks,''
\newblock in {\em International Conference on Frontiers in Handwriting Recognition}, 2016

\bibitem{HMMs}
Urs-Viktor Marti and Horst Bunke
\newblock ``Using a Statistical Language Model to Improve the Performance of an HMM-Based Cursive Handwriting Recognition Systems,''
\newblock in  {\em International Journal of Pattern Recognition and Artificial Intelligence}, 2001

\bibitem{Doe}
Patrick Doetsch and Michał Kozielski and Hermann Ney 
\newblock ``Fast and Robust Training of Recurrent Neural Networks for Offline Handwriting Recognition,''
\newblock in {\em International Conference on Frontiers in Handwriting Recognition}, 2014.

\bibitem{hmm1}
Horst Bunke and M. Roth and E.G. Schukat-Talamazzini
\newblock ``Off-line Cursive Handwriting Recognition using Hidden Markov Models,''
\newblock in {\em  IEE Proceedings - Vision, Image and Signal Processing}, 1995.

\bibitem{hmm2}
Radmilo {Bozinovic} and Sargur {Srihari}
\newblock ``Off-line cursive script word recognition,''
\newblock in {\em  IEEE Transactions on Pattern Analysis and Machine Intelligence}, 1989

\bibitem{GramCTC}
Hairong Liu and Zhenyao Zhu and  Xiangang Li and Sanjeev Satheesh
\newblock ``Gram-CTC: Automatic Unit Selection and Target Decomposition for Sequence,''
\newblock in {\em International Conference on Machine Learning}, 2017

\bibitem{LatSeqDe}
 William Chan and Yu Zhang and Quoc Le and Navdeep Jaitly
\newblock ``Latent Sequence Decompositions,''
\newblock in {\em International Conference on Learning Representations}, 2017

\bibitem{Sanab2}
Thomas Zenkel and Ramon Sanabria and Florian Metze and Alex Waibel
\newblock ``Subword and Crossword Units for {CTC} Acoustic Models,''
\newblock in {\em Interspeech}, 2018


\bibitem{s2s}
Arindam Chowdhury and Lovekesh Vig
\newblock ``An Efficient End-to-End Neural Model for Handwritten Text Recognition,''
\newblock in {\em British Machine Vision Conference}, 2018

\bibitem{castro}
Dayvid Castro and Byron L. D. Bezerra and Meuser Valenca
\newblock ``Boosting the Deep Multidimensional Long-Short-Term Memory Network for Handwritten Recognition Systems,''
\newblock in {\em International Conference on Frontiers in Handwriting Recognition}, 2018

\bibitem{harald}
Harald Scheidl and Stefan Fiel and Robert Sablatnig
\newblock ``Word Beam Search: A Connectionist Temporal Classification Decoding Algorithm,''
\newblock in {\em International Conference on Frontiers in Handwriting Recognition}, 2018

\bibitem{hmtlother}
Kalpesh Krishna and Shubham Toshniwal and Karen Livescu
\newblock ``Hierarchical Multitask Learning for CTC-based Speech Recognition,''
\newblock in {\em CoRR}, abs/1807.06234, 2018

\bibitem{s2shtr}
Johannes Michael and Roger Labahn and Tobias Gr{\"{u}}ning and Jochen Z{\"{o}}llner
\newblock ``Evaluating Sequence-to-Sequence Models for Handwritten Text Recognition,''
\newblock in {\em CoRR}, abs/1903.07377, 2019


\end{thebibliography}
\end{document}